\documentclass[conference,a4paper]{IEEEtran}
\IEEEoverridecommandlockouts

\usepackage[hidelinks]{hyperref}
\usepackage[cmex10]{amsmath}
\usepackage{amssymb,amsfonts}
\usepackage{dblfloatfix}

\usepackage[ruled,vlined]{algorithm2e}
\usepackage{graphicx}
\graphicspath{{Figures/PDF/}{Figures/PNG/}}

\usepackage{booktabs}
\usepackage{siunitx}
\usepackage[numbers,compress]{natbib}
\usepackage{texnames}
\usepackage{bm,bbm}
\usepackage{orcidlink}

\newcommand{\vect}[1]{\boldsymbol{#1}}
\newcommand{\mat}[1]{\boldsymbol{\uppercase{#1}}}
\newcommand{\ten}[1]{\textbf{$\mathcal{\uppercase{#1}}$}}

\begin{document}

\title{Do Tensorized Large-Scale Spatiotemporal Dynamic
Atmospheric Data Exhibit Low-Rank Properties?
}

\author{	\IEEEauthorblockN{Ryan Solgi\orcidlink{0000-0002-7560-5210}\thanks{Accepted to IEEE IGARSS 2025}\\}
	\IEEEauthorblockA{\textit{University of California-Satna Barbara}\\
		California, USA\\
		solgi@ucsb.edu}
	\and
	\IEEEauthorblockN{Seyedali Mousavinezhad}
	\IEEEauthorblockA{\textit{University of Texas at Austin}\\
		Texas, USA\\
		}
	\and
	\IEEEauthorblockN{Hugo A. Loaiciga}
	\IEEEauthorblockA{\textit{University of California-Satna Barbara}\\
		California, USA\\
		}
	
}

\maketitle
\begin{abstract}
	In this study, we investigate for the first time the low-rank properties of a tensorized large-scale spatio-temporal dynamic atmospheric variable. We focus on the Sentinel-5P tropospheric NO2 product (S5P-TN) over a four-year period in an area that encompasses the contiguous United States (CONUS). Here, it is demonstrated that a low-rank approximation of such a dynamic variable is feasible. We apply the low-rank properties of the S5P-TN data to inpaint gaps in the Sentinel-5P product by adopting a low-rank tensor model (LRTM) based on the CANDECOMP / PARAFAC (CP) decomposition and alternating least squares (ALS). Furthermore, we evaluate the LRTM's results by comparing them with spatial interpolation using geostatistics, and conduct a comprehensive spatial statistical and temporal analysis of the S5P-TN product. The results of this study demonstrated that the tensor completion successfully reconstructs the missing values in the S5P-TN product, particularly in the presence of extended cloud obscuration, predicting outliers and identifying hotspots, when the data is tensorized over extended spatial and temporal scales. 
\end{abstract}

\begin{IEEEkeywords}
	Tensor Decomposition, Low Rank, Remote Sensing, Spatiotemporal, Atmospheric Data 
\end{IEEEkeywords}

\section{Introduction}

Estimating missing values in remote sensing data has been a long-standing problem of study~\cite{Chen_Savitzky_2004, Shen_review_2015}. Research in this domain generally falls into two categories: data fusion, where missing values are predicted by integrating information from auxiliary sources~\cite{Li_SAR_fusion_2024, Xiang_SAR_fusion_2024}, and inpainting, where gaps are self-reconstructed using only the available original data~\cite{Miao_dictionary_2019,Wang_landsat_2021, Wang_unsupervised_2023}. In both categories, various methods including regressions~\cite{Rulloni_gap_2012}, geostatistics~\cite{Zhang_kriging_2007}, deep learning~\cite{Li_cnn_2017}, and low-rank~\cite{Fellah_lowrank_2017} models among others~\cite{Wang_radiance_2019} were studied with extensive applications in land surface monitoring including optical images~\cite{Wang_image_2022} and normalized difference vegetation index (NDVI)~\cite{Li_ndvi_review_2021}.

Atmospheric data often exhibit more dynamic behavior than land surface observations~\cite{Zhu_DecSolNet_2021}, making its recovery particularly challenging in the presence of large temporal or spatial gaps. Recovery of gaps in various atmospheric observations is an emerging topic, and several recent studies have addressed this issue~\cite{Lops_no2_2023, Bai_tucker_2022, Betancourt_ozone_2023, Zhu_DecSolNet_2021, Zhang_aerosol_2022}. However, all of these methods fall into the category of data fusion to predict missing values and rely on auxiliary data such as ground-level measurements, location information, meteorological parameters or assimilation models. However, the literature on inpainting methods for recovering missing gaps in atmospheric observations is currently limited, usually focusing on simple approaches, including computing the mean of the data~\cite{Wei_review_2019} or using interpolations and geostatistical techniques, including kriging~\cite{Yu_Kriging_2011}.

Several studies have applied spatial and temporal dependencies simultaneously to improve the accuracy of data estimation~\cite{Chen_regression_2017, Zhang_dcnn_2018, Chu_ndvi_2022, Lops_no2_2023, Yeganeh_no2_spatiotemporal_2018}. Previous spatiotemporal methods have primarily been used for land-surface data imputation or fusion-based atmospheric data recovery, and the literature on spatiotemporal analysis for atmospheric data inpainting is scarce.

Several studies have advanced tensor completion algorithms for optical remote sensing image denoising and cloud removal~\cite{Ng_weighted_tensor_2017, He_tensor_ring_2019, Ji_sparse_reg_2022, Zhang_tt_2024, Zou_deep_prior_2024}. Spatiotemporal tensor modeling was applied for the long-term recovery of NDVI time series~\cite{Chu_tensor_ndvi_2021}. In particular in atmospheric data, Tucker's decomposition was implemented to fuse various data to reconstruct the aerosol optical depth (AOD) data in China~\cite{Bai_tucker_2022}. Despite recent advances in tensor completion for recovery of land surface observations, its application for atmospheric data inpainting has never been reported. This is primarily because small-scale atmospheric data are often not temporally stationary and may not exhibit low-rank properties on small spatial scales. 

To the best of our knowledge, a comprehensive analysis of the low-rank properties of large-scale spatiotemporal atmospheric data has not been reported in the literature. This raises an important question: Do large-scale spatiotemporal dynamic atmospheric data exhibit low-rank properties? Answering this question has at least one significant implication: that inpainting of atmospheric data based on low-rank properties may be feasible. Therefore, we investigate for the first time whether tensorized large-scale spatiotemporal atmospheric observations exhibit low-rank properties. We focus on the tropospheric NO\textsubscript{2} product of Sentinel-5P over a four-year period in the contiguous United States (CONUS). Here we organize satellite point data into spatiotemporal tensors at different resolutions and adopt a low-rank tensor model (LRTM) based on CP decomposition and alternating least squares (ALS) to complete the NO\textsubscript{2} data. We thoroughly assess the accuracy of data recovery for both random and cloud-pattern missing values, which resemble the actual conditions of the original data in both patterns and coverage, using a cloud pattern transfer masking (CPTM) technique. Our examination also includes statistical testing of the datasets and comparison with Kriging. The results provide new insights into such atmospheric data.
\section{Methods}
\subsection{Background}
Given a general 3-way tensor $\ten{X} \in \mathbb{R}^{I_1 \times I_2 \times I_3}$, let set $\boldsymbol{\Omega}=\{[i_1,i_2,i_3]|\mat{X}_{i_1}[i_2,i_3]\neq \varnothing, 1\leq i_k \leq I_k, k=1,2,3 \}$ represent the indices of all observed entries in the tensor $\ten{X}$, where $\mat{X}_{i_1}$ denotes a slice of the tensor $\ten{X}$ along its first dimension. Let $\mathcal{P} \in \{0,1\}^{I_1 \times I_2 \times I_3}$ be a projection tensor corresponding to the observed entries of $\ten{X}$ defined as follows:

\begin{align}
\label{eq: projection}
\ten{P}[i_1,i_2,i_3]=\begin{cases}
1,&{\text{if}}\ [i_1,i_2,i_3] \in \boldsymbol{\Omega},\\ 
{0,}&{\text{if}}\ [i_1,i_2,i_3] \notin \boldsymbol{\Omega}.
\end{cases}
\end{align}

The projection of the error between the original and recovered tensors is minimized as follows:

\begin{align}
\label{eq:TC}
\min_{\mat{A}_1,\mat{A}_2,\mat{A}_3} \Vert (\ten{X}-\ten{\hat{X}})\odot \ten{P}\Vert_{F}^2,
\end{align}
where $\ten{\hat{x}}\equiv[\boldsymbol{\lambda};\mat{A}_{1},\mat{A}_{2},\mat{A}_{3}]$ refers to the CP decomposition in Kruskal format~\cite{minster_cp_2022}, $\mat{A}_i\in \mathbb{R}^{I_i\times R}$, $R$ denotes the CP rank, and operator $\odot$ denotes hadamard (elementwise) product. Using alternating least squares (ALS) Eq.~\eqref{eq:TC} is divided into $\sum_{l=1}^{3}I_l$ independent least-squares problems that are solved iteratively for $T$ iterations~\cite{minster_cp_2022}. Solving each least square problem updates a row of $\mat{A}_i=(\vect{\alpha}_1,\cdots,\vect{\alpha}_{I_i})^T$ for $i=1,2,3$, as follows:

\begin{align}
\label{eq: closed-form}
\vect{\alpha}_j=(\mat{G}_i\odot(\vect{p}_j \otimes \vect{1}_R))^{-1} (\vect{x}_j\odot \vect{p}_j), \quad \forall j=1,\cdots,I_i,
\end{align}
where $\vect{x}_j=(\mat{X}_{(i)}[j,1],\cdots,\mat{X}_{(i)}[j,m])^T$, $\vect{p}_j=(\mat{P}_{(i)}[j,1],\cdots,\mat{P}_{(i)}[j,m])^T$, $m=\prod_{l=1, l\neq i}^{3}I_l$, $\mat{G}_i$ represents column-wise Kronecker (also known as Khatri-Rao) product of all factor matrices except $\mat{A}_i$ \cite{Liu_tensor_2022}, the operators $\otimes$ denotes tensor product, $\vect{1}_R$ denotes a vector of ones of size $R$, $\mat{Q}^{-1}$ denotes the inverse of matrix $\mat{Q}$, $\mat{X}_{(i)}$ and $\mat{P}_{(i)}$ denotes unfolded tensor $\ten{X}$ and $\ten{P}$ along $i^{th}$ dimension, respectively.

After computing $[\boldsymbol{\lambda};\mat{A}_{1},\mat{A}_{2},\mat{A}_{3}]$, a missing entry at position $i_1$, $i_2$, and $i_3$ is estimated as follows:

\begin{align}
\label{eq:recover}
\ten{\hat{X}}[i_1,i_2,i_3]=\sum_{r=1}^{R} (\lambda_r \prod_{l=1}^{3}\mat{A}_l[i_l,r]).
\end{align}

\subsection{Tensorizataion}
The Sentinel-5P products consist of scattered point data, with coordinates that do not necessarily align from one day to the next. In this work, the daily point data are first resampled into a daily raster without applying interpolation to organize the data into a tensor. We start by intersecting the point data with a spatial bounding box $\boldsymbol{\Theta}=\{(\phi,\psi)|\phi_{min}\leq\phi\leq\phi_{max}, \psi_{min}\leq \psi \leq \psi_{max}\}$, which covers the extend of study area. Here $\phi$ and $\psi$ represent latitude and longitude, respectively. Next the low-quality data are filtered out if their quality assurance ($q_a$) scores are below the recommended threshold. The remaining high-quality point data are then rasterized into pixels. In this study, we apply a constrained nearest neighbor method for resampling. If no point measurement fell onto a pixel, the pixel is left unassigned and considered missing. Otherwise, it is assigned the value of the nearest point to the grid. Next, the daily rasters are concatenated to construct a 3-way tensor of Sentinel-5P product denoted by $\ten{X}\in \mathbb{R}^{I_1 \times I_2 \times I_3}$ where $I_1=D$ represents the temporal dimension, and $I_2=\left\lceil\frac{\phi_{max}-\phi_{min}}{\delta}\right\rceil$ and $I_3=\left\lceil\frac{\psi_{max}-\psi_{min}}{\delta}\right\rceil$ correspond to the latitudinal and longitudinal dimensions, respectively. Given the tensorized product $\ten{X}$ with partially missing entries, our objective is to predict the missing values by developing a low-rank model based on the observed entries, assuming that the tensor $\ten{x}$ is low-rank. We studied the rasterization of point data at various resolutions including $\delta=0.5^{\circ}$ and $\delta=0.25^{\circ}$. Given the excessive time complexity of Kriging for the size of this data, we evaluated the LRTM algorithm against Kriging at the coarse resolution of $0.5^{\circ}$ and further analyzed the LRTM method at the finer resolution of $0.25^{\circ}$.

\subsection{Complementary Statistical Analysis}

To determine whether the data is spatially stationary and exhibit spatial dependencies a semivariogram, is applied to the temporally averaged data. To study temporal dependencies and assess whether the data is temporally stationary, we spatially average the data and analyze the resulting time series. Two statistical tests are applied to the times series. The Augmented Dickey-Fuller (ADF) test~\cite{ADF_1981} is used to test the null hypothesis that a unit root is present in the time series, with the alternative hypothesis that the data is stationary. Additionally, the Ljung-Box (LB) test~\cite{Ljung_Box_1978} is applied to test the null hypothesis that the time series is independently distributed, with the alternative hypothesis that it is not independently distributed.

\subsection{Additional Synthetic Missing Values}

In this study, we add synthetic missing values to the data in addition to those already present in the Sentinel-5P product. This approach allows the evaluation of the performance of the methods presented in this study under two different scenarios: randomly generated missing values and cloud pattern missing values. we develop synthetic cloud-like structured missing regions to accurately reflect the actual gaps in the product that resemble cloud patterns. To closely simulate these actual missing regions, we applied a technique referred to here as cloud pattern transfer masking (CPTM). The CPTM generates a mask from all missing regions from one randomly selected day (source) and apply the same mask to another randomly selected day (target) that has no gaps in the region of the synthetic mask. Therefore, a synthetic day is generated that combines all actual missing regions from both the source and target days while retaining the remaining good quality data from the target day. 
\subsection{Metrics}
To evaluate the results we apply two metrics including correlation (r) and index of agreement (IOA) computed as follows:

\begin{align}
    r=\frac{\sum_{i}(p_i-\bar{p})(o_i-\bar{o})}{\sqrt{\sum_i(p_i-\bar{p})^2\sum_i(o_i-\bar{o})^2}}\\
    IOA=1-\frac{\sum_i(p_i-o_i)^2}{\sum_i(\lvert p_i-\bar{o}\rvert+\lvert o_i-\bar{o}\rvert)^2}
\end{align}
where $p_i$ and $o_i$ are the $i$-th prediction and observation, respectively, and $\bar{p}$ and $\bar{o}$ are the means of predictions and observations, respectively.
\section{Case Study}
\begin{figure}
\centering
\includegraphics[width=3.5 in]{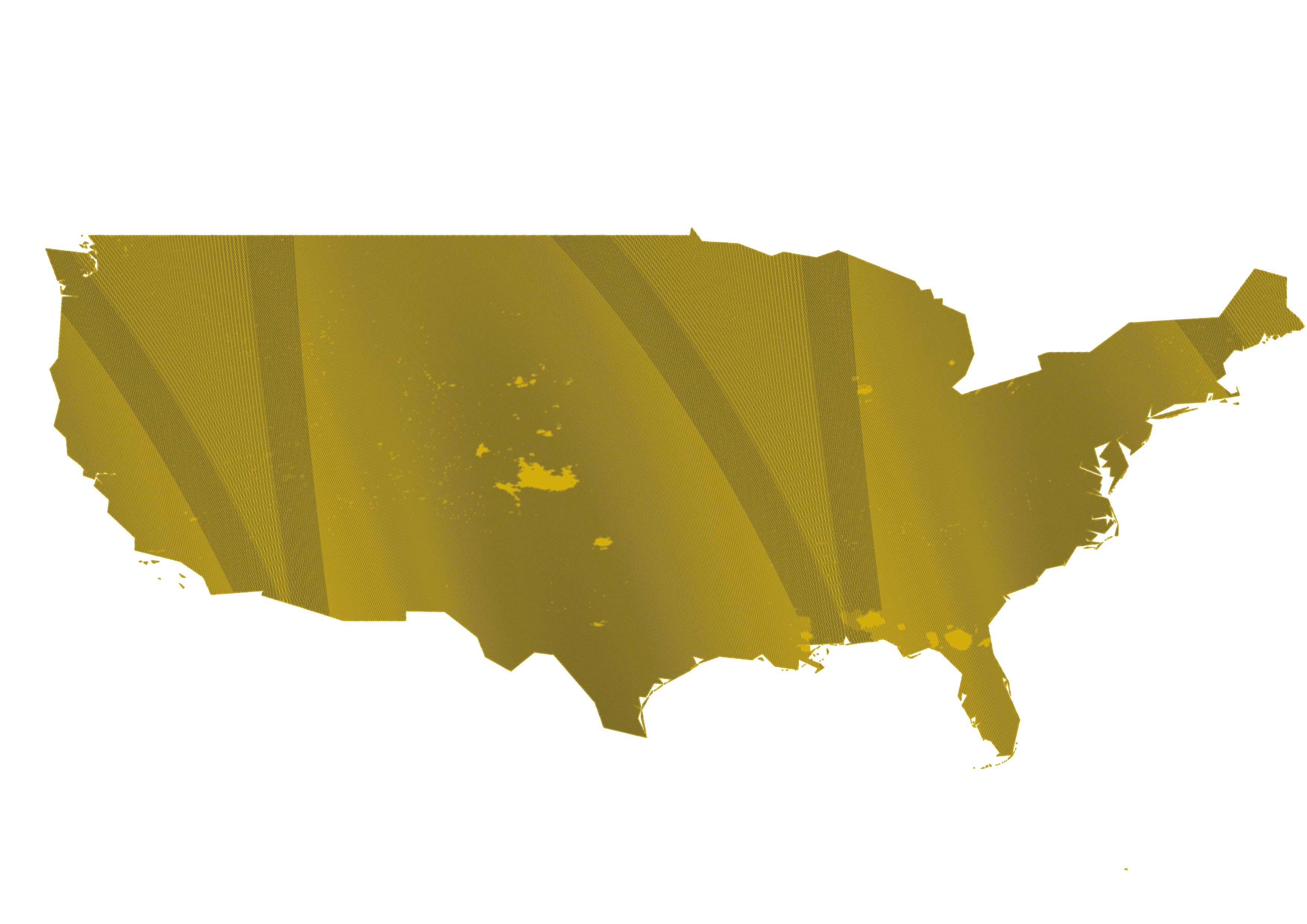}
\caption{Locations of Measurements with $q_a > 0.5$ (good quality data)(black circles) of S5P-TN data collected on 06/15/2023 over CONUS (yellow background).}
\label{fig:sample-measurements}
\end{figure}

\begin{figure}
\centering
\includegraphics[width=3.5in]{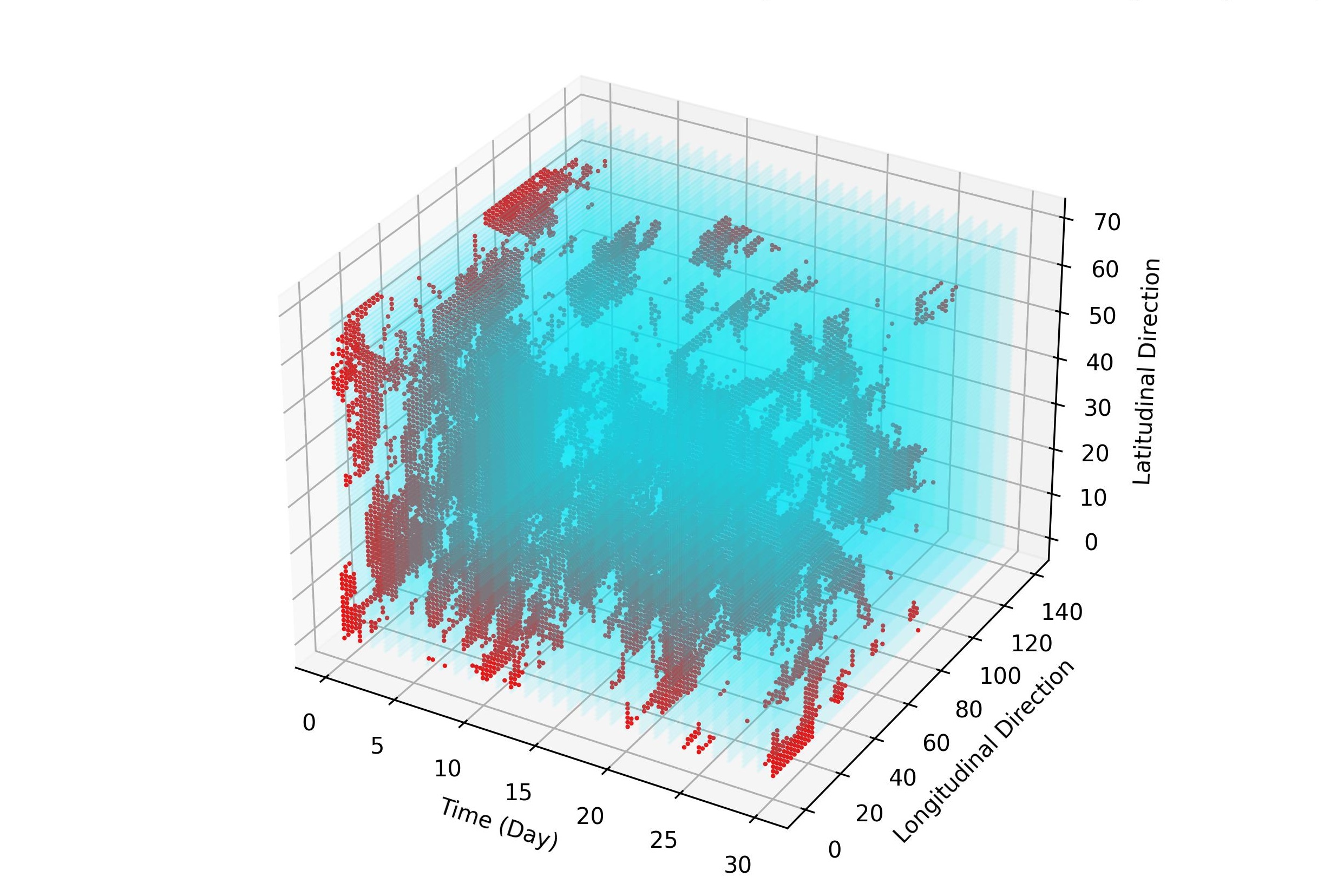}
\caption{Three-dimensional visualization of the spatial and temporal positions of high-quality S5P-TN data (blue dots) and missing positions (red dots) for January 2019 in the study area.}
\label{fig:3d}
\end{figure}

This work investigates the low-rank properties of the tensorized NO\textsubscript{2} product from Sentinel-5P over an area bounded by $20^{\circ}N$ to $55^{\circ}N$ latitude and $60^{\circ}W$ to $130^{\circ}W$ longitude, encompassing the CONUS, for a four-year period from January 1, 2019, to January 1, 2023. The Sentinel-5P tropospheric NO\textsubscript{2} (S5P-TN) product is one of the most widely used products of Sentinel-5P~\cite{Geffen_no2_2022}. The S5P-TN data is accessible through NASA's Goddard Earth Sciences Data and Information Services Center (GES DISC) \cite{copernicus_sentinel_data_processed_by_esa_koninklijk_nederlands_meteorologisch_instituut_knmi_sentinel-5p_2018}. The initial data provided by GES DISC consists of point data with daily global coverage. The S5P-TN data are associated with quality assurance ($q_a$) values ranging from zero to one, with higher values indicating greater confidence in the data. The recommended minimum $q_a$ value is 0.5~\cite{eskes_sentinel-5_2022}. Fig.~\ref{fig:sample-measurements}(b) shows the locations where S5P-TN data with $q_a > 0.5$ were collected over CONUS on June 15, 2023. In a spatial coverage area such as CONUS, low-quality data resulting in missing regions are encountered almost daily. Consequently, dealing with missing values is inevitable when working with this data. Fig.~\ref{fig:3d} illustrates an example of the missing values in the tensorized S5P-TN data for the study area in January 2019 with 5.41\% missing values. The initial unit of the S5P-TN product was $mol/m^2$. Since $molec/cm^2$ unit is more commonly used than $mol/m^2$, the values were converted into $molec/cm^2$ ($1 molec/cm^2 = 6.02214\times 10^{19} mol/m^2$) for pixels with valid values. Finally the data were normalized to fall withing the range $[0,1]$. The normalized data were used for the analysis. After tensorization of the data, the tensor dimensions (day, latitude, longitude) or (depth, height, width) for $0.5^{\circ}$ and $0.25^{\circ}$ resolutions are $(1461,70,140)$, $(1461,141,281)$, respectively. At the $0.5^{\circ}$ resolution, only 88 days had no missing values, while at the $0.25^{\circ}$ resolution, only one day out of 1,461 days had no missing values. Additionally, we observe that the majority of days at both resolutions have less than 10\% missing values. Overall, the dataset contains 2.3\% and 3.0\% missing values for the $0.5^{\circ}$ and $0.25^{\circ}$ resolutions, respectively.

\section{Results}

\subsection{Statistical Spatial and Temporal Analysis}

For temporally averaged tensorized S5P-TN data across the domain of study at $0.5^{\circ}$ resolution, the exponential theoretical semivariogram best fitted the experimental semivariogram with a correlation coefficient of 0.85. Since there was a good fit between experimental and theoretical semivariograms, and also the semivariogram reached a plateau, the intrinsic hypothesis holds and the data is spatially stationary.

For the spatially averaged normalized S5P-TN data from 1/1/2019 to 1/1/2023. The ADF statics was -4.93 and with a p-value of $3\times 10^{-5}$ (the critical value at 1\% is -3.435), the null hypothesis is rejected with high confidence. With a LB test statistics of 1378.16 and a p-value $1.6\times 10^{-298}$, the null hypothesis is rejected with a high confidence. Therefore, the data is not temporally independent and does not exhibit non-stationarity, as both null hypotheses were rejected.

Given these results, analyzing the data spatially and temporally simultaneously may yield better predictions than using only spatial interpolation. Therefore, LRTM, which enables simultaneous spatiotemporal analysis, is a promising method for imputing the missing values in the S5P-TN data.

\subsection{Added Random Missing Values}
Given the tensorized S5P-TN has about 3\% missing values in total, 3\% extra random missing values were added to the tensorized S5P-TN product and used the added missing values to study LRTM and compare it with Kriging. The remaining 94\% data were used to train Kriging and derive low-rank factors in LRTM. Table~\ref{table:performance-050} compares the performance of LRTM and Kriging at $0.5^{\circ}$ resolution and compares LRTM at $0.25^{\circ}$ resolution with the results of inverse distance weighting (IDW) and depthwise partial convolutional neural networks (DW-PCNN) from a previous study~\cite{Lops_no2_2023} that fused data assimilation with S5P-TN product for spatiotemporal estimation of S5P-TN missing values at a finer resolution. For $0.5^{\circ}$ and $0.25^{\circ}$ resolutions, the best LRTM results are achieved at ranks 500 and 1,800, respectively, although even lower ranks outperformed other methods including kriging. It is notable that LRTM, using only the original data without any auxiliary information, achieved competitive accuracy compared to a spatiotemporal fusion model (i.e., DW-PCNN) that integrated assimilation data with the original data to predict missing values.

\begin{table}
\caption{Performance of LRTM comparing to Kriging and previous studies for randomly missing values at different resolutions.}
\label{table:performance-050}
\begin{center}
\begin{tabular}{c|ccc}
\multicolumn{1}{c}{Method}
&\multicolumn{1}{c}{IOA}
&\multicolumn{1}{c}{r}
&\multicolumn{1}{c}{MAE}
\\ \hline


\textbf{LRTM} ($\delta=0.5^{\circ}$) & \textbf{0.88} & \textbf{0.80} & \textbf{0.00379}\\
Kriging & 0.73 & 0.65 & 0.00428\\
\hline
\textbf{LRTM} ($\delta=0.25^{\circ}$) & \textbf{0.91} & \textbf{0.84} & \textbf{0.001569}\\
IDW (2019)\textsuperscript{*}  & 0.82 & 0.69 & -\\
DW-PCNN (2019)\textsuperscript{*} & 0.85 & 0.76 & -\\
DW-PCNN (2020)\textsuperscript{*} & 0.88 & 0.79& -\\
\hline
\end{tabular}
\par \vspace{0.2cm}
\footnotesize{Note: Values with * are cited from a previous study~\cite{Lops_no2_2023}, and the year in parentheses indicates the portion of the data used from that specific year.}
\end{center}
\end{table}

\subsection{Synthetic Cloud Pattern Prediction}
Using the CPTM, we generated 300 synthetic days with cloud pattern missing regions. Table~\ref{table:cloud_metrics} lists the metrics for synthetic cloud pattern prediction using both LRTM and Kriging. Compared to Kriging, LRTM demonstrated significant improvements, with IOA increasing by more than 0.18 and correlation by over 0.14.  

\begin{table}
\caption{Performance of LRTM and Kriging for added synthetic cloud missing values at $0.5^{\circ}$ and $0.25^{\circ}$ resolutions.}
\label{table:cloud_metrics}
\begin{center}
\begin{tabular}{cc|ccc}
\multicolumn{1}{c}{Resolution}
&\multicolumn{1}{c}{Method}
&\multicolumn{1}{c}{IOA}
&\multicolumn{1}{c}{r}
&\multicolumn{1}{c}{MAE}
\\ \hline
$0.5^{\circ}$ & Kriging & 0.64 & 0.54 & 0.0052\\
$0.5^{\circ}$ & LRTM & 0.82 & 0.69 & 0.0056\\
\hline
$0.25^{\circ}$ & LRTM & 0.80 & 0.67 & 0.0022\\
\hline
\end{tabular}
\end{center}
\end{table}

\begin{figure}
    \centering
    \includegraphics[width=3.5in]{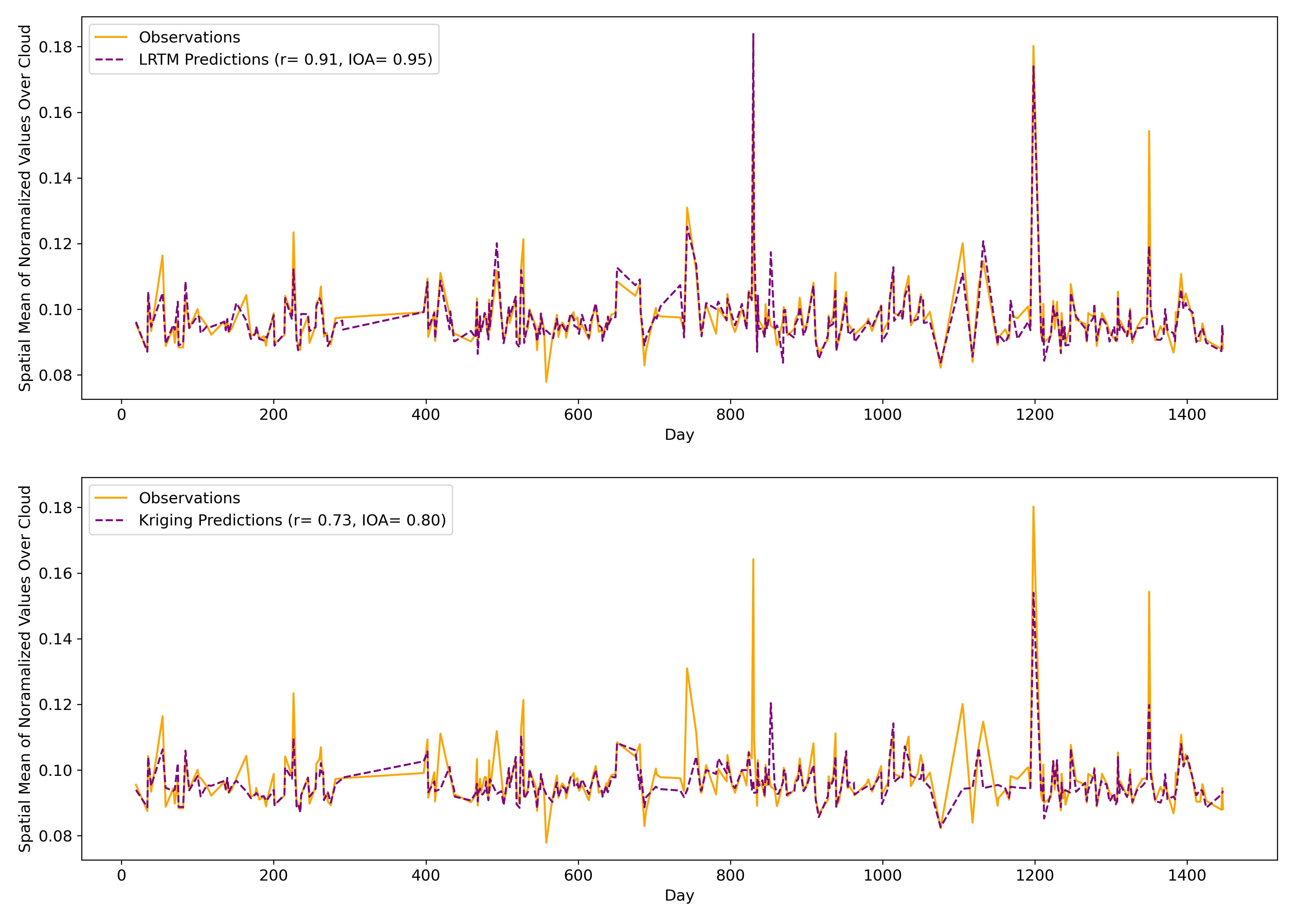}   
    \caption{Spatial mean of normalized values over synthetic cloud region on different days.}
    \label{fig:spatial-mean-syn-cloud}
\end{figure}

Fig.~\ref{fig:spatial-mean-syn-cloud} illustrates the spatially averaged predictions and observations across the entire cloud regions per day at $0.5^{\circ}$. On average, the correlation (r) and IOA for LRTM were 0.91 and 0.95, for $0.5^{\circ}$ and $0.25^{\circ}$, respectively, while for Kriging, they were 0.73 and 0.80, representing significant differences of approximately 0.18 and 0.15, respectively.

\begin{figure}
    \centering
    \includegraphics[width=3.5in]{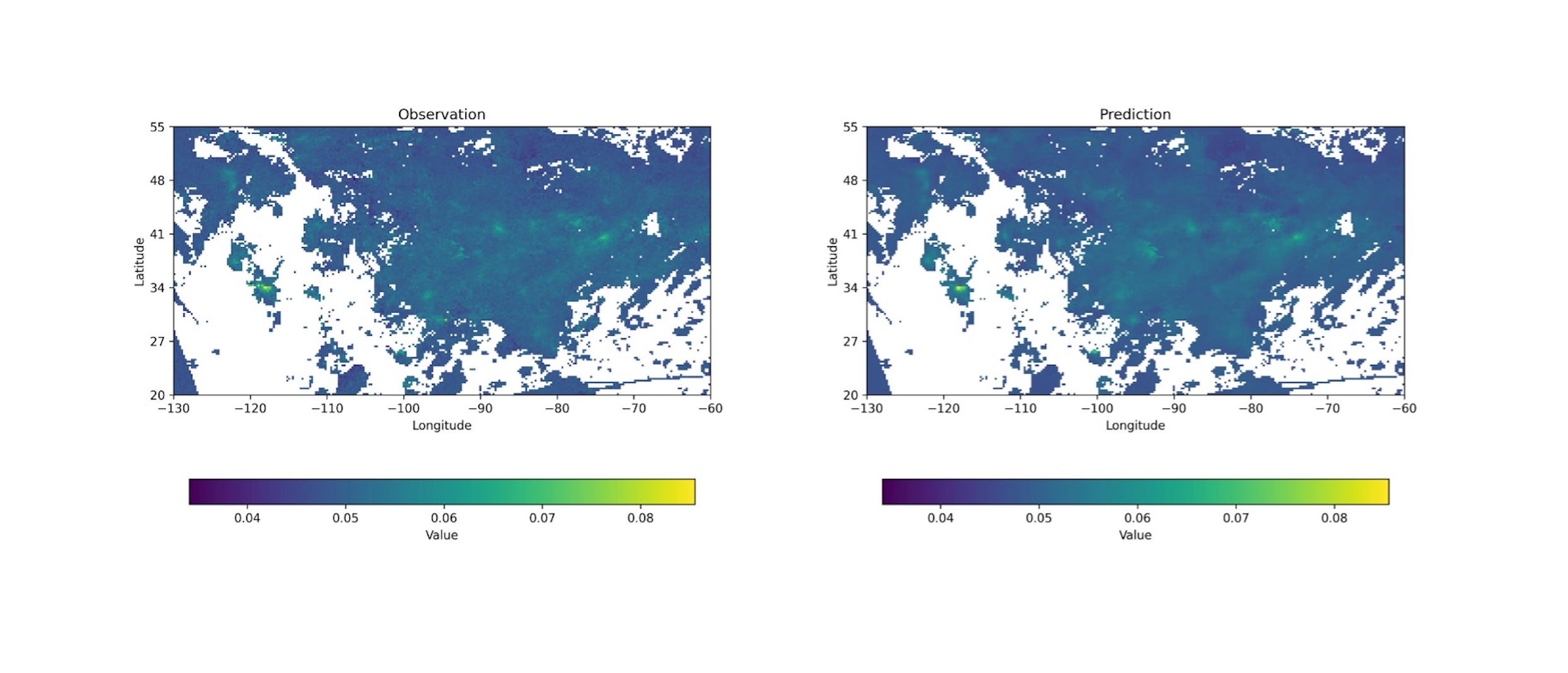}   
    \caption{Observations and predictions of S5P-TN data for the collection of all synthetic cloud regions at $0.25^{\circ}$ resolution (white color presents regions with no synthetic cloud).}
    \label{fig:cloud_collection}
\end{figure}

Fig.~\ref{fig:cloud_collection} visualizes the synthetic cloud pattern predictions versus observations by aggregating all synthetic cloud regions at a $0.25^{\circ}$ resolution. A notable observation in Fig.~\ref{fig:cloud_collection} is that LRTM accurately predicted the overall pattern of spatial NO\textsubscript{2} fluctuations across the study area, including pollution hotspots. One prominent example is seen in the cloud region above the Los Angeles area, with latitude and longitude roughly equal to 118°W (-118) and 34°N, respectively.

\section{Conclusion}
This study applied a low-rank tensor model (LRTM) to factorize and reconstruct the tensorized tropospheric NO\textsubscript{2} product from Sentinal-5P over a four-year period across the CONUS. The accuracy of data recovery of the random and cloud-pattern missing values was thoroughly examined and the results were compared with that of geostatistics. The superior results of LRTM for imputing the added random missing values and the cloud pattern missing values demonstrated that a low-rank approximation of large-scale dynamic atmospheric products is possible when we tensorize the data across extended temporal and spatial scales, simultaneously.

\bibliographystyle{IEEEtranN}
\bibliography{main}

\end{document}